\pdfoutput=1

\documentclass[11pt,a4paper]{article}
\usepackage{times,latexsym}
\usepackage{url}
\usepackage[T1]{fontenc}

\usepackage[hyperref,acceptedwithA]{tacl2021v1}

\usepackage{xspace,mfirstuc,tabulary}

\newif\iftaclinstructions
\taclinstructionsfalse %
\iftaclinstructions
\renewcommand{\confidential}{}
\renewcommand{\anonsubtext}{(No author info supplied here, for consistency with
TACL-submission anonymization requirements)}
\newcommand{\instr}
\fi
\usepackage{graphicx}
\usepackage{booktabs}
\usepackage{csquotes}
\usepackage{amsmath}
\usepackage{placeins}
\iftaclpubformat %

\else

\fi

\usepackage{enumitem}

\title{Leaving the Cave by Breaking the Symmetry:\\Identifiable Structure Can Enable Cross-Lingual Generalization}

\newcommand{\addonly}{\textsc{add-only}}
\newcommand{\subonly}{\textsc{sub-only}}
\newcommand{\addmax}{\textsc{add+max}}
\newcommand{\addsub}{\textsc{add+sub}}
\newcommand{\addcopy}{\textsc{add+add-copy}}

\usepackage{tikz}
\usetikzlibrary{calc,arrows.meta,positioning}
\usepackage{graphicx}
\usepackage{subcaption}
\definecolor{lang0col}{RGB}{234,95,148}
\definecolor{lang1col}{RGB}{0,0,255}
\definecolor{linkcol}{RGB}{74,65,168}

\tikzset{
  vbox/.style={rounded corners=2pt, draw, inner sep=0pt, align=center,
               font=\scriptsize, text width=17mm, minimum height=13mm},
  l0/.style={draw=lang0col, fill=lang0col!8},
  l1/.style={draw=lang1col, fill=lang1col!8},
  sh/.style={draw=black!55, fill=black!5, rounded corners=2.5pt, align=center,
             minimum width=17mm, minimum height=20mm, font=\footnotesize},
  ar/.style={-{Latex[length=1.6mm]}, draw=black!55, line width=0.6pt},
  ring0/.style={draw=lang0col, line width=1pt},
  ring1/.style={draw=lang1col, line width=1pt},
  plane/.style={draw=black!45, dashed, rounded corners=3pt,
                minimum width=72mm, minimum height=30mm},
  lk/.style={linkcol, densely dashed, line width=0.7pt},
  lkb/.style={linkcol!60, densely dashed, line width=0.7pt},
}

\newcommand{\panelA}{%
\begin{tikzpicture}
  \node[vbox,l0] (in0) at (0,0.72) {%
    \begin{tabular}{@{}r@{\;}c@{\;}l@{}}
      3 & $+$ & 42\\ 13 & $+$ & 16\\ 100 & $+$ & 20
    \end{tabular}};
  \node[vbox,l1] (in1) at (0,-0.72) {%
    \begin{tabular}{@{}l@{\;}c@{\;}l@{}}
      III & $\boxplus$ & XLII\\ XIII & $\boxplus$ & XVI\\ C & $\boxplus$ & XX
    \end{tabular}};
  \node[sh] (tf) at (3.7,0) {shared\\Transformer};
  \node[font=\tiny, text=black!55, below=0.4mm of tf] {$\bmod\ 113$};
  \node[vbox,l0] (out0) at (7.4,0.72) {%
    \begin{tabular}{@{}c@{}} 45\\ 29\\ 7 \end{tabular}};
  \node[vbox,l1] (out1) at (7.4,-0.72) {%
    \begin{tabular}{@{}c@{}} XLV\\ XXIX\\ VII \end{tabular}};
  \draw[ar] (in0.east)  -- (in0.east  -| tf.west);
  \draw[ar] (in1.east)  -- (in1.east  -| tf.west);
  \draw[ar] (out0.west -| tf.east) -- (out0.west);
  \draw[ar] (out1.west -| tf.east) -- (out1.west);
\end{tikzpicture}}

\newcommand{\clockticks}[2]{%
  \foreach \a in {0,30,...,330}{\fill[#2] ($(#1)+(\a:0.80)$) circle (1.3pt);}}
 
\newcommand{\panelB}{%
\begin{tikzpicture}
  \node[plane] (pl) {};
\node[
  anchor=north,
  font=\tiny,
  text=black!55,
  align=center
] at (pl.north) {Shared answer subspace\\
                 Corresponding numbers unaligned};
  \coordinate (c0) at (-1.6,-0.25);
  \coordinate (c1) at ( 1.6,-0.25);
  \draw[ring0] (c0) circle (0.72); \clockticks{c0}{lang0col!55}
  \draw[ring1] (c1) circle (0.72); \clockticks{c1}{lang1col!55}
  \coordinate (A42) at ($(c0)+(300:0.80)$);
  \coordinate (B42) at ($(c1)+(90:0.80)$);
  \coordinate (A3)  at ($(c0)+(30:0.80)$);
  \coordinate (B3)  at ($(c1)+(250:0.80)$);
  \draw[lk]  (A42) -- (B42);
  \draw[lkb] (A3)  -- (B3);
  \foreach \p in {A42,B42,A3,B3}{\fill[white] (\p) circle (2.4pt);}
  \fill[linkcol]    (A42) circle (2.0pt) (B42) circle (2.0pt);
  \fill[linkcol!65] (A3)  circle (2.0pt) (B3)  circle (2.0pt);
  \node[font=\tiny, text=linkcol]    at ($(c0)+(300:1.00)$) {42};
  \node[font=\tiny, text=linkcol]    at ($(c1)+(90:1.06)$)  {XLII};
  \node[font=\tiny, text=linkcol!70] at ($(c0)+(30:1.02)$)  {3};
  \node[font=\tiny, text=linkcol!70] at ($(c1)+(250:1.00)$) {III};
\end{tikzpicture}}
 
\newcommand{\panelC}{%
\begin{tikzpicture}
  \node[plane] (pl) {};
\node[
  anchor=north,
  font=\tiny,
  text=black!55,
  align=center
] at (pl.north) {Shared answer subspace\\
                 Corresponding numbers aligned};
  \coordinate (c0) at (-1.6,-0.25);
  \coordinate (c1) at ( 1.6,-0.25);
  \draw[ring0] (c0) circle (0.72); \clockticks{c0}{lang0col!55}
  \draw[ring1] (c1) circle (0.72); \clockticks{c1}{lang1col!55}
  \coordinate (A42) at ($(c0)+(90:0.80)$);
  \coordinate (B42) at ($(c1)+(90:0.80)$);
  \coordinate (A3)  at ($(c0)+(270:0.80)$);
  \coordinate (B3)  at ($(c1)+(270:0.80)$);
  \draw[lk]  (A42) -- (B42);
  \draw[lkb] (A3)  -- (B3);
  \foreach \p in {A42,B42,A3,B3}{\fill[white] (\p) circle (2.4pt);}
  \fill[linkcol]    (A42) circle (2.0pt) (B42) circle (2.0pt);
  \fill[linkcol!65] (A3)  circle (2.0pt) (B3)  circle (2.0pt);
  \node[font=\tiny, text=linkcol]    at ($(c0)+(90:1.00)$)  {42};
  \node[font=\tiny, text=linkcol]    at ($(c1)+(90:1.06)$)  {XLII};
  \node[font=\tiny, text=linkcol!70] at ($(c0)+(270:1.00)$) {3};
  \node[font=\tiny, text=linkcol!70] at ($(c1)+(270:1.00)$) {III};
\end{tikzpicture}}

\newcommand{\sd}[1]{\ {\scriptsize$\pm #1$}}
\usepackage{cleveref}
\usepackage[table]{xcolor}

\newcommand{\focal}[1]{%
  \begingroup
  \setlength{\fboxsep}{2pt}%
  \colorbox{orange!30}{\textbf{#1}}%
  \endgroup
}
\usepackage{amsmath}
\usepackage{amssymb}

\crefname{figure}{Fig.}{Figs.}
\Crefname{figure}{Fig.}{Figs.}

\crefname{table}{Tab.}{Tabs.}
\Crefname{table}{Tab.}{Tabs.}

\crefname{section}{Sec.}{Secs.}
\Crefname{section}{Sec.}{Secs.}

\crefname{subsection}{Sec.}{Secs.}
\Crefname{subsection}{Sec.}{Secs.}

\crefname{appendix}{App.}{Apps.}
\Crefname{appendix}{App.}{Apps.}

\crefname{equation}{Eq.}{Eqs.}
\Crefname{equation}{Eq.}{Eqs.}

\author{
  Frederick Riemenschneider
  \and
  Anette Frank
  \\
  Department of Computational Linguistics
  \\
  Heidelberg University, Germany
  \\
  \texttt{\{riemenschneider|frank\}@cl.uni-heidelberg.de}
}

\date{}
\begin{document}
\maketitle
\begin{abstract}
Why do multilingual language models generalize across languages even without shared vocabulary? The Platonic Representation Hypothesis posits that models trained on different data, objectives, and modalities converge toward a shared statistical model of reality. Extending this intuition to representations inside a single multilingual model, we propose the \textit{Relational Identifiability Hypothesis}: relations in a shared world can promote entity alignment by ruling out alternative cross-lingual correspondences, making corresponding entities more likely to become directly readable without an additional transformation. We test this hypothesis in a controlled bilingual Transformer whose two token-disjoint languages describe the same modular-arithmetic world.
We find that the languages can develop a shared answer subspace without aligning corresponding numbers. Cross-lingual alignment emerges selectively when the available relations constrain number correspondences: subtraction makes zero task-level identifiable, inducing selective alignment, while an order-sensitive operation produces broad alignment across numbers. These patterns predict success in zero-shot cross-lingual transfer. Together, our results show that relational structure in a shared world can promote entity alignment and enable transfer.

\end{abstract}

\section{Introduction}
For almost a decade, research on cross-lingual generalization has
been in explanatory disarray: Why do multilingual language models (MLLMs) generalize across languages? One line of work partly attributes the cross-lingual abilities of MLLMs to shared subwords, which could anchor alignment across languages \citep[][\textit{inter alia}]{pires-etal-2019-multilingual,
wu-dredze-2019-beto,Cao2020Multilingual,lin-etal-2019-choosing}. Other studies,
by contrast, find that cross-lingual generalization persists even when lexical
anchors are absent \citep[][\textit{inter alia}]{ketal,dufter-schutze-2020-identifying,schaeferetal,kallini-etal-2025-false}.  In the extreme, this has motivated researchers to deliberately reduce lexical overlap between languages during MLLM training \citep{liang-etal-2023-xlm}.

In a different domain, the \emph{Platonic Representation Hypothesis} (PRH)
proposes that \enquote{neural networks, trained with different objectives
on different data and modalities, are converging to a shared statistical model of reality in their representation spaces}
\citep{pmlr-v235-huh24a}. The name invokes Plato's idea that the objects
we encounter are imperfect manifestations of a deeper underlying reality:
different models may likewise arrive at compatible representations because
they are constrained by the same world.
This idea has not been developed in the context of MLLMs. However, languages can be viewed as different modalities for describing a shared world \citep{parcalabescu-etal-2021-multimodality}. Their vocabularies, grammatical structures, and cultural perspectives differ, but they refer to many of the same objects, events,
properties, and relations in our world.

A direct extension of the PRH to language would ask whether monolingual models trained separately on different languages converge toward compatible representations. Since
such models inhabit separate representation spaces, establishing correspondence between them requires
a transformation or a coordinate-invariant comparison. We study a stronger form of alignment within a single multilingual model: we ask whether corresponding entities in two languages with {completely disjoint vocabularies} can align in a single, shared 
model’s existing representational space, \emph{without an additional cross-lingual transformation}. 

We frame this question through the \emph{Relational Identifiability Hypothesis}: relations in a shared world can promote cross-lingual entity alignment by reducing the ambiguity about which entities correspond across languages. An entity should be more likely to acquire an aligned representation when the remaining alternatives can no longer change its cross-lingual identity, whereas entities left undetermined may remain unaligned even within a shared model. In our controlled setting, we make this notion exact by analyzing which input-output relabelings preserve the available arithmetic tasks. An answer is \emph{task-level iden\-ti\-fi\-able} if every such relabeling leaves its identity unchanged. This condition removes one source of ambiguity, but does not guarantee that train\-ing will realize the resulting alignment perfectly.

\begin{figure}[t]
  \centering
  \begin{subfigure}{.9\linewidth}\centering
    \resizebox{\linewidth}{!}{\panelA}
    \caption{\textbf{Setup.} A shared Transformer is trained jointly on two languages with disjoint input and output vocabularies.}    \label{fig:concept-a}
  \end{subfigure}\\[5pt]
  \begin{subfigure}{.9\linewidth}\centering
    \resizebox{\linewidth}{!}{\panelB}
    \caption{\textbf{\addonly{}.} The answer representations occupy closely overlapping subspaces, but corresponding numerical answers are not aligned across languages (cosine: $0.015$).}
    \label{fig:concept-b}
  \end{subfigure}\\[5pt]
  \begin{subfigure}{.9\linewidth}\centering
    \resizebox{\linewidth}{!}{\panelC}
    \caption{\textbf{\addmax{}.} Adding an order-sen\-si\-tive operation makes the directions for corresponding answers substantially more similar
(cosine: $0.843$).}
    \label{fig:concept-c}
  \end{subfigure}
  \caption{
Modular addition produces a shared answer subspace without number-level alignment. Adding an order-sen\-si\-tive operation broadly aligns those numbers despite disjoint vocabularies.}
  \label{fig:concept}
\end{figure}

Starting from the modular addition setup of \citet{nanda2023progress}, we construct a controlled bilingual setting in which two token-disjoint languages describe the same mathematical world. We systematically vary the operations available during training and measure how strongly the induced relational structure constrains possible cross-lingual correspondences (see \Cref{fig:concept}).
We find that a shared answer subspace does not by itself align corresponding
numbers. When models are trained only on modular addition, the two languages develop closely overlapping answer subspaces, yet the corresponding numbers remain unreadable across languages. Adding other operations changes this pattern: subtraction makes the number zero identifiable and produces selective alignment, whereas the \(\max\) operation produces
broad alignment across numbers.

The behavioral consequences of this 
distinction are mirrored in {zero-shot cross-lingual transfer} -- the canonical test for cross-lingual generalization. We treat the models trained under the different arithmetic conditions as pre-trained models and ask which downstream properties transfer cross-lingually without target-language supervision. The results follow the same pattern: properties {that involve task-identifiable, aligned answers transfer substantially
better across languages.
Our contributions are threefold:
\begin{enumerate}[
  label=\roman*),
  noitemsep,
  leftmargin=*,
  labelsep=.25em,
  align=left
]\item We introduce the \emph{Relational Identifiability Hypothesis}: relational constraints can promote cross-lingual entity
alignment by ruling out alternative correspondences. To our knowledge, we are
the first to formulate and mechanistically isolate relational identifiability
as a source of cross-lingual alignment. 
\item We develop a controlled mechanistic setting to test this hypothesis. We
show that a shared, causally relevant answer subspace can emerge without
number-level alignment. The operations available during training determine
which answer identities are fixed at the task level, and learned alignment
patterns reflect these answer-specific distinctions.
\item We demonstrate that these answer-specific
alignment patterns predict which downstream properties transfer
cross-lingually.
\end{enumerate}

\section{Related Work}

\paragraph{Cross-Lingual Generalization.}
Explanations of cross-lingual generalization have focused on both \emph{linguistic
factors}, including language relatedness, word order, and shared structure, and
\emph{architectural factors}, such as model capacity, depth, and parameter sharing
\citep[][\textit{inter alia}]{lin-etal-2019-choosing,lauscher-etal-2020-zero,
dufter-schutze-2020-identifying,deshpande-etal-2022-bert,ketal}.
The role of lexical overlap has been particularly contested. Shared subwords
can facilitate cross-lingual representational alignment and
transfer
\citep[][\textit{inter alia}]{pires-etal-2019-multilingual,
wu-dredze-2019-beto,Cao2020Multilingual,kallini-etal-2025-false},
but cross-lingual generalization can persist when vocabulary overlap is reduced or removed
\citep[][\textit{inter alia}]{ketal,dufter-schutze-2020-identifying,conneau-etal-2020-emerging,deshpande-etal-2022-bert}.

Recent work has moved beyond identifying factors associated with transfer
to examining the \emph{representations} that support it. For example,
 \citet{wendler-etal-2024-llamas}  ask whether 
\textsc{Llama} \citep{touvron2023llama} uses English as an internal
pivot. \citet{dumas-etal-2025-separating} show that MLLMs represent language and conceptual content separately. %
Finally, \citet{riemenschneider-frank-2025-cross} document how language-specific representations develop into cross-lingually shared abstractions over the course of pre-training. Together, these findings support the existence of cross-lingually shared representations. However, natural multilingual data contain many potential sources of alignment simultaneously (lexical overlap, parallel data, code-switching, and other statistical regularities). This makes it difficult to isolate the factors that enable cross-lingual generalization \enquote{in the wild}.

A separate methodological tradition therefore uses controlled training settings to isolate such factors more directly. Studies based on transformed vocabularies and synthetic languages manipulate the information shared across languages and show that transfer can survive substantial reductions in lexical overlap and other surface cues \citep{dufter-schutze-2020-identifying,ketal,deshpande-etal-2022-bert,schaeferetal,kallini-etal-2025-false}. 
Our work lies at the extreme end of this controlled tradition. We replace natural-language data with a fully specified mathematical world in which the entities, relations, and correct computations are known by construction. This allows us to remove lexical overlap entirely, vary the amount and nature of relational structures available during training, and directly test which numbers align across two languages in diverse settings.

\paragraph{Representational Convergence.}

The PRH proposes that models trained on
different data, objectives, and modalities can converge toward compatible
representations of a shared underlying world
\citep{pmlr-v235-huh24a}. Subsequent work has provided both constructive and
theoretical support for this possibility. \citet{jha2025harnessing}
show that different text encoders learn sufficiently similar embedding geometry to permit translation without paired examples. In the restricted setting of embedded deep linear networks,
\citet{ziyin2025proof} prove that networks of different widths and
depths trained on different data can learn equivalent representations up to a
rotation.

These studies examine \textit{compatibility} across separately learned 
models or 
representation spaces. We, instead, dissociate compatibility of representational spaces from 
\textit{alignment} within a \textit{single} and \textit{bilingual model}, in which two languages use completely disjoint vocabularies but share the Transformer body. This setting allows us to ask whether shared relational structure is
sufficient to induce cross-lingual alignment and under which conditions such
alignment emerges.

\paragraph{Mechanistic Studies of Arithmetic.}
Our experimental setup builds on the modular-addition task of
\citet{nanda2023progress}, who reverse-engineer the algorithm learned
by a small Transformer model. The model represents numbers using Fourier features
and implements modular addition through trigonometric operations (cf. \Cref{sec:toymodel}). Closely
related structures have since been identified in pre-trained language models:
\citet{zhou2024pretrained} find that Fourier features play a causal role in
addition, while \citet{kantamneni2025language} identify causally relevant helical representations of numbers and propose that language models perform addition by manipulating these representations to construct the answer representation.
We use this mechanistically tractable setting to study %
{under which conditions} two token-disjoint languages develop cross-lingually aligned representations of the same mathematical entities as the operations available during training change.

\section{Multilingual Modular Arithmetic}
\label{sec:toymodel}
We build on a well-understood toy Transformer, which we adapt into a \enquote{multilingual} variant. \citet{nanda2023progress} studied a small Transformer trained on modular addition: the model receives two inputs $a,b\in\{0,\dots,p-1\}$ for some prime $p$ and must predict their sum $c=a+b \bmod p$. Although trained on only \(30\%\) of the input
pairs, the model ultimately generalizes perfectly to the held-out \(70\%\), a
phenomenon known as \emph{grokking}.

Importantly for our purposes, the learned mechanism in this standard setup has been reverse-engineered in detail.  \citet{nanda2023progress} show that the model solves modular addition using the so-called \textit{Clock} algorithm \citep{zhong2023the}.\footnote{Depending on the specific setup and architecture, other algorithms may emerge \citep{zhong2023the}; we focus here on the standard model that 
implements the canonical solution.}

Intuitively, the {Clock} algorithm mirrors how the hours on a clock face wrap around: the model represents each number as an angle on a circle and performs addition by rotating along it.
Since the circle closes on itself after a full turn, this rotation automatically yields the sum modulo $p$.

More concretely, the {Clock} algorithm operates in parallel over a small set of \emph{key frequencies} \(w_k=\frac{2k\pi}{p}\). At each frequency, the embedding represents \(a\) and \(b\) by their sine and cosine coordinates on a circle. The attention and MLP layers combine these coordinates using angle-addition identities to represent \(a+b\), and the output layer compares the resulting positions with each candidate answer \(c\). Summing across key frequencies produces constructive interference at \(c=a+b\bmod p\) and destructive interference elsewhere. %

We extend this setup to a {multilingual} setting. Each language has its own input and output
tokens, while both share the Transformer body. The languages are trained
jointly in a single model and optimization run. Thus, they describe the same
mathematical entities and operations through different lexical interfaces,
with no overlap in their vocabularies.
This provides a controlled testbed for examining shared computational
and representational structure, task-level identifiability, and learned
cross-lingual alignment.

The advantage and novelty of our toy setting is that it provides a fully controlled, minimal, and idealized environment for mechanistic studies of cross-lingual generalization, with well-understood circuits.
Such 
analyses would not be possible with real language data and models at scale.

\subsection{Training Conditions}
\label{sec:conditions}

The Relational Identifiability Hypothesis predicts that cross-lingual
alignment should vary with which number identities are fixed by the
operations available during training. We therefore compare conditions that
impose different relational constraints while holding the languages,
architecture, and numerical domain fixed.

We distinguish the following conditions:
\vspace{-.5\baselineskip}

\begin{description}[noitemsep]
\item[\addonly{}.] Both languages express modular addition. This is the minimal bilingual
extension of the original modular-addition setup.
\item[\subonly{}.] The model learns modular subtraction in both languages.
Subtraction still permits ambiguity among nonzero numbers, but it uniquely
distinguishes zero through equations of the form \(a-a=0\).

\item[\addsub{}.] The model learns both modular addition and modular subtraction in each
language. It tests whether adding subtraction to the addition-only setting changes which numbers become identifiable across languages.

\item[\addmax{}.] The model jointly learns addition 
and \(\max(a,b)\), which adds information about ordinary numerical order and thus more strongly constrains how individual
numbers can correspond across languages.

\item[\addcopy{}.] The model learns a second copy of modular addition, distinguished by a separate task prefix. This setting serves as a control for whether alignment can arise merely from additional task supervision.

\end{description}

\paragraph{Training Details.}
All five conditions use the same bilingual Transformer
with disjoint language vocabularies over \(p=113\) numbers.
Each input begins
with a language-specific task prefix, 
while tasks within a language share number tokens. We use the same randomly selected \(30\%\) of all
\(p^2\) operand pairs across languages and, in the multitask conditions, across tasks. Models are evaluated on the remaining \(70\%\).

We train a one-layer, four-headed Transformer with a 128-dimensional residual
stream and a 512-dimensional MLP for 50,000 training steps, using the complete
training set at each step. For every condition, we analyze 20 independently
initialized models that achieve at least \(98\%\) held-out accuracy in every
active task-language combination, and report means and standard deviations
across models.

Across all five conditions, Fourier ablations confirm that the models rely on
task-appropriate Fourier-structured causal mechanisms: ablating the selected
Fourier directions reduces task accuracy by at least \(0.961\) in every model.
Complete training parameters and condition-specific pass rates are provided in
App.~\ref{app:training-details}, and full Fourier-validation results are
reported in \Cref{tab:appendix-fourier-validation}.

\subsection{Which Numbers Do Tasks Determine?}
\label{sec:math}

The two languages describe the same numbers, but the training data provide no direct cross-lingual links between corresponding number tokens. To characterize this ambiguity, we ask whether the numerical
interpretation of one language can be systematically relabeled while preserving the arithmetic task.

Because our models use separate parameters for input and output number tokens, these two roles may be relabeled differently. Let
\(\sigma,\tau:\mathbb{Z}_p\rightarrow\mathbb{Z}_p\) be permutations, where
\(\sigma\) relabels numbers when they occur as operands and
\(\tau\) relabels numbers when they occur as answers. We call a pair
\((\sigma,\tau)\) \emph{task-preserving} for an operation \(f\) if
\begin{equation}
\tau\bigl(f(a,b)\bigr)
=
f\bigl(\sigma(a),\sigma(b)\bigr)
\qquad
\text{for all } a,b\in\mathbb{Z}_p.
\label{eq:permutations}
\end{equation}
Under such a relabeling, every equation remains correct although token identities may have changed. For modular addition, these task-preserving relabelings are
\begin{equation}
\sigma(x)=kx+\delta,
\qquad
\tau(x)=kx+2\delta,
\label{eq:sandtau}
\end{equation}
where \(k\in\mathbb{Z}_p\setminus\{0\}\) and
\(\delta\in\mathbb{Z}_p\). The identity
\[
(ka+\delta)+(kb+\delta)
=
k(a+b)+2\delta
\] 
verifies that these maps preserve addition. Conversely, setting \(b=0\) in \Cref{eq:permutations} gives \(\tau(a)=\sigma(a)+\sigma(0)\). Substituting this expression back shows that \(x\mapsto\sigma(x)-\sigma(0)\) is additive and hence has the form \(kx\) over \(\mathbb{Z}_p\). Therefore, no other task-preserving relabelings exist.

The offset occurs twice on the answer side because an addition answer combines two operands. For \(p=113\),
addition therefore permits \(p(p-1)=12{,}656\) task-preserving relabelings.
Varying \(\delta\) can move every answer, including zero, since \(p\) is odd
and \(2\delta\) ranges over all of \(\mathbb{Z}_p\). Thus, modular addition alone determines no answer token's identity.

The available relabelings depend on the task.
For subtraction, 
the
operand offset cancels,
\[
(ka+\delta)-(kb+\delta)=k(a-b),
\]
so the answer-side relabeling is \(\tau(x)=kx\). Every such relabeling 
satisfies \(\tau(0)=0\), whereas nonzero numbers can still be rearranged by varying \(k\).
Repeating addition under a second task prefix in \addcopy{} adds the same constraint again and removes no ambiguity. Combining addition with subtraction in \addsub{}
removes the offset \(\delta\), but retains the multiplicative relabelings.

For \(\max\), setting \(a=b\) in \Cref{eq:permutations} gives
\(\tau(a)=\sigma(a)\). The resulting shared relabeling must preserve the
ordinary numerical order on \(\{0,\ldots,p-1\}\). The only order-preserving
permutation of this finite ordered set is the identity. Thus, combining \(\max\) with addition in \addmax{} eliminates every non-identity task-preserving relabeling.

\begin{table}[t]
\centering
\small
\setlength{\tabcolsep}{4pt}
\begin{tabular}{@{}lrr@{}}
\toprule
\textbf{Training Condition}
&
\textbf{Relabelings}
&
\textbf{Fixed by All} \\
\midrule
\addonly{}
&
\(p(p-1)\)
&
none \\
\addcopy{}
&
\(p(p-1)\)
&
none \\
\subonly{}
&
\(p(p-1)\)
&
\(0\) \\
\addsub{}
&
\(p-1\)
&
\(0\) \\
\addmax{}
&
\(1\)
&
all \\
\bottomrule
\end{tabular}

\caption{
Task-level predictions for 
all conditions.
An answer is fixed 
if
every task-pre\-ser\-ving output relabeling
\(\tau\) maps it to itself. For \(p=113\),
\(p(p-1)=12{,}656\) and \(p-1=112\).
}
\label{tab:relabelings}
\end{table}

We say that an answer \(c\) is \emph{task-level identifiable} when every task-preserving
relabeling \((\sigma,\tau)\) fixes it on the answer side, that is, when \(\tau(c)=c\). If some task-preserving output relabeling moves an answer, correct task performance alone cannot force that answer to have the intended
cross-lingual identity. If every such relabeling fixes it, the operation removes this particular source of ambiguity.
This analysis clarifies what identities the task determines, but leaves open the internal organization learned by the model. In particular, it does not imply that the two languages inside a trained model must be related by any task-preserving relabeling, nor does task-level identifiability guarantee perfectly learned alignment.

\subsection{Evaluating Cross-Lingual Alignment}
We use three complementary diagnostics of cross-lingual alignment and
representational sharing: i) whether answer representations are directly
readable across languages, ii) whether corresponding answers point in similar
directions in the residual stream, and iii) whether answer information lies in the same low-dimensional subspace.

\paragraph{Cross-Readout.}
We test whether the languages are \emph{directly mutually readable}. For each
direction \(\ell_s \rightarrow \ell_t\), we run source-language inputs through the model and restrict the logits to the target language's \(p\) output-number tokens. Cross-readout is
correct when the highest-scoring target-language token denotes the
ground-truth answer to the source-language arithmetic problem.

\paragraph{Cosine Similarity.}
We compare the hidden states associated with corresponding numerical answers in the two languages. For each language, we collect the final residual-stream activation at the equals position and average it over all examples whose ground-truth answer is \(c\), yielding one answer signature for each \(c\in\{0,\ldots,p-1\}\). We center these signatures within each language, compute the cosine similarity between signatures for corresponding answers, and average the resulting similarities over \(c\). High cosine similarity indicates that corresponding answers are encoded along similar directions in the residual stream.

\paragraph{Subspace Similarity.}
Cosine similarity requires corresponding answers to point in similar directions. The languages could nevertheless encode individual answers differently within the same low-dimensional residual-stream subspace. To test this, we form each language's \(p \times d_{\mathrm{model}}\) matrix of answer signatures and center it by subtracting the mean across answers. Singular value decomposition yields the principal residual-stream directions along which the answer signatures vary. For a fixed rank \(r\), the leading \(r\) directions define an answer subspace for each language. The
answer representations have an effective rank of roughly \(8\)--\(10\),
consistent with the low-dimensional Fourier structure of the original modular
addition algorithm. We therefore use \(r=8\) throughout; results are similar for nearby ranks (\Cref{tab:appendix-rank-sensitivity}). We measure their alignment using the mean of the cosines of the canonical angles between the two subspaces.

Even unrelated low-rank subspaces can exhibit nonzero overlap by chance. We
therefore compare the observed overlap with that of randomly drawn subspaces
of the same rank and ambient dimension. Moreover, successful models may
generally represent answers in similar low-dimensional subspaces for reasons
unrelated to cross-lingual interaction. We therefore also compare the
within-model cross-language overlap with the same-language overlap between
independently trained bilingual models. For each condition, we form ten
disjoint pairs from the 20 frozen models and average the two same-language
comparisons within each pair.

This tests a broader form of cross-lingual representational sharing than
matched-answer cosine similarity: individual answers may point in different directions while the answer representations nevertheless lie within the same subspace.

\paragraph{Subspace Ablation.}
Subspace overlap is observational and does not show that the shared directions are actually used by the model. We therefore test their causal relevance by removing one
language's rank-\(r\) answer subspace from the residual stream while evaluating the \emph{other} language. 

For each language \(\ell\) and intervention site \(s\), we use training
examples to estimate a mean activation
\(\boldsymbol{\mu}_{\ell,s}\) and an orthonormal rank-\(r\) answer-subspace
basis \(\mathbf{B}_{\ell,s}\). Let \(\ell\) denote the language being
evaluated and \(\ell'\) the other language. To remove the directions identified
from the other language, we replace the evaluated language's equals-position
activation \(\mathbf{x}\) with
\[
\mathbf{x}'
=
\mathbf{x}
-
\mathbf{B}_{\ell',s}^{}\mathbf{B}_{\ell',s}^{\top}
\bigl(\mathbf{x}-\boldsymbol{\mu}_{\ell,s}^{}\bigr).
\]
This removes the component of the evaluated language's activation that lies
in the other language's answer subspace while preserving the evaluated
language's mean activation. We use \(r=8\) and intervene either immediately
before the MLP or at the final residual stream. We define the ablation effect
as the difference between the model's original and ablated held-out accuracy.

We compare the other-language ablation with three controls. The
\emph{own-language} intervention uses
\(\mathbf{B}_{\ell,s}\) and serves as a positive control. The
\emph{random-subspace} control removes an isotropically sampled subspace of the
same rank. Finally, the \emph{independent-model} control removes the
same-language answer subspace estimated from a separately initialized model
while still evaluating the original model. This tests whether the effect is
specific to the coordinate system shared within one bilingual model, rather
than being a generic consequence of removing directions learned by any model
that solves the task. Complete intervention definitions are provided in App.~\ref{app:subspace-ablation}, and the full ablation results and controls are reported in \Cref{tab:appendix-subspace-ablation}.

\section{Shared Structure Without Alignment}
We first examine \addonly{}, for which the task-level analysis predicts that
no answer identity is determined. Modular addition permits task-preserving
relabelings that move every answer, so correct performance does not require
the two languages to agree on which number each answer representation denotes.
This condition therefore tests whether a shared model can develop common
structure without aligning corresponding entities. All models generalize nearly perfectly and rely on task-appropriate Fourier components (cf.\ App.\ \ref{app:fourier-validation}). 

\paragraph{Corresponding Numbers Remain Unaligned.} As predicted, the two languages show little identity alignment
(\Cref{tab:symmetry-main}, col. 1). Cross-readout accuracy is only
\(0.011\), barely above chance (\(\frac{1}{113} \approx 0.009\)), and the mean cosine similarity between answer signatures for corresponding numbers is \(0.015\).
Thus, an answer representation produced in one language is not directly readable through the other language's output vocabulary and does not point in the same direction as the representation of the corresponding answer in that language.

\paragraph{Shared Structure Remains.}

Despite the absence of identity alignment, the answer representations occupy
strongly overlapping subspaces. At rank \(8\), cross-language overlap is \(0.944\), compared with \(0.216\) for independently trained bilingual models and \(0.211\) for random subspaces (App.~\Cref{tab:appendix-subspace-overlap-controls}).
These subspaces are also causally relevant across languages. Removing the answer subspace estimated from the other language reduces
held-out addition accuracy by \(0.989\) at the final residual stream. By
contrast, subspaces estimated from independently trained models reduce
accuracy by only \(0.001\) at either intervention site. The two languages
therefore rely on strongly overlapping, model-specific answer subspaces that
are causally relevant across languages, even though corresponding numbers
occupy different directions within them. These results show that strongly overlapping answer structure can emerge even when the task leaves number identities
undetermined. As predicted by the task-level analysis, such shared structure
does not by itself force corresponding numbers to align.

\begin{table*}[t]
\centering

\small
\setlength{\tabcolsep}{4pt}
\begin{tabular}{@{}lccccc@{}}
\toprule
\textbf{Metric} & \textbf{\addonly{}} & \textbf{\addcopy{}} & \textbf{\subonly{}} & \textbf{\addsub{}} &  \textbf{\addmax{}} \\
\cmidrule(r){1-2} \cmidrule(l){3-6}
\multicolumn{6}{@{}l}{\textbf{Broad Identity Alignment}} \\
\quad Identity Cross-Readout & $.011$\sd{.008} & $.012$\sd{.007} &  $\hphantom{-}.018$\sd{.014} & $.012$\sd{.007} & \focal{$.125$\sd{.075}} \\
\quad Matched-Answer Cosine & $.015$\sd{.074} & $.003$\sd{.014} & $-.008$\sd{.026} & $.039$\sd{.125} &  \focal{$.843$\sd{.221}} \\
\cmidrule(r){1-2} \cmidrule(l){3-6}
\multicolumn{6}{@{}l}{\textbf{Shared Causal Answer Subspace}} \\
\quad Rank-8 Answer-Subspace Overlap & $.944$\sd{.080} & $.928$\sd{.091} & $\hphantom{-}.889$\sd{.086} & $.855$\sd{.176} & $.924 $\sd{.086} \\
\quad Cross-Ablation Drop, Pre-MLP & $.928$\sd{.036} & $.923$\sd{.046} & $\hphantom{-}.956$\sd{.046} & $.947$\sd{.047} &  $.983$\sd{.005} \\
\quad Cross-Ablation Drop, Final & $.989 $\sd{.006} & $.985$\sd{.017} & $\hphantom{-}.988$\sd{.009} & $.935$\sd{.127} &  $.979$\sd{.010} \\
\bottomrule
\end{tabular}
\caption{Identity alignment and shared causal answer structure across
20 bilingual models per condition. Cells report means
$\pm$ standard deviations across models. Broad identity scores average the 113 answer-specific scores within each model. Identity cross-readout uses held-out operand pairs and the other language's 113 output-number tokens. Subspace bases are fitted on training examples, ablation drops use held-out examples. The focal task is addition except for \subonly{}, where it is subtraction.}
\label{tab:symmetry-main}
\end{table*}

\section{Breaking the Symmetry}
\label{sec:breaking}

The \addonly{} results show that shared representational structure is not
sufficient to align corresponding numbers: the two languages rely on closely
overlapping answer subspaces, but do not agree on which
directions denote which numbers.
The relational structure induced by modular addition alone does not determine
the identities of the answer tokens across languages. Multiple
task-preserving relabelings remain possible, each compatible with correct
addition. We refer to this remaining ambiguity as a \emph{symmetry} of the
task. The model may therefore develop strongly overlapping answer structure
while assigning corresponding numbers to different directions within it.

The real world, however, supplies many more relational constraints. Entities
participate in multiple relations simultaneously, so a candidate cross-lingual correspondence must be compatible with all of them. Additional relations can break the task symmetry by ruling out alternatives left open by
a single operation. Our {Relational Identifiability Hypothesis} proposes that corresponding entities are more likely to align when the available relations rule out alternatives that change their identities. We test this prediction in the remaining conditions, which differ in the number identities determined by the available operations.

\paragraph{Alignment Reflects Task-Level Identifiability.}
The 
five conditions produce sharply different %
cross-lingual
alignment patterns (\Cref{tab:symmetry-main}). 
As expected, duplicated addition changes little: \addcopy{} closely resembles \addonly{}, with near-chance cross-readout and near-zero cosine despite strongly overlapping answer subspaces.
Hence, additional samples and a second task prefix are not sufficient to align numbers.

Broad averages for \subonly{} and \addsub{} likewise show little overall
alignment.
Under \addsub{}, identity cross-readout is \(0.012\);
matched-answer cosine is \(0.039\).
Yet, these averages %
conceal an answer-specific effect
(\Cref{tab:symmetry-answer-breakdown}): under \addsub{}, answer-zero
cross-readout rises to \(0.658\) and cosine similarity to \(0.792\), while nonzero answers remain largely unaligned. The \subonly{} condition reproduces this selectivity even more strongly, with answer-zero cross-readout of \(0.775\) and cosine similarity of \(0.907\). Subtraction therefore selectively aligns zero, the one answer whose identity
it determines, rather than aligning the numerical system broadly. Direction-level analyses are reported in App.~\ref{app:alignment-details}.

By contrast, \addmax{} produces broad directional alignment.
Strikingly, matched-answer cosine rises from \(0.015\) under \addonly{}
training to \(0.843\) under \addmax{}. Corresponding answers therefore occupy
similar residual-stream directions once training also requires an
order-sensitive operation. Identity cross-readout also rises above its
near-chance level under addition, but remains much lower at \(0.125\). This gap shows that directional alignment is only one component of direct
cross-language readability. We conclude that introducing an order-sensitive
operation that eliminates the remaining task-preserving alternatives promotes
broad directional alignment, even though the two languages {do not share any tokens}.

\paragraph{The Shared Subspace Persists.}
The conditions differ sharply in identity alignment, but much less in the
degree to which the two languages share an answer subspace
(\Cref{tab:symmetry-main}). Rank-8 answer-subspace overlap remains high across all five conditions, ranging from \(0.855\) to \(0.944\). Cross-language ablations also produce large accuracy drops at both intervention sites in every condition. Thus, the two languages continue to rely on closely overlapping answer subspaces even when corresponding numbers remain misaligned. 
Taken together, the results show that cross-lingual alignment is neither
automatic nor all-or-none. Although shared answer structure persists across
conditions, the identities aligned within it vary with the available
relational constraints. This pattern supports the Relational Identifiability
Hypothesis: corresponding entities are more likely to align when the available relations {rule out alternatives that change their identities}.

\paragraph{Tying Embeddings and Unembeddings Aligns Zero.}
Task-level ambiguities may also relate to model architecture. In our  models, separate input embeddings and output unembeddings allow
operand and answer tokens to be relabeled differently. For addition, this
permits the offset in
\(\sigma(x)=kx+\delta\) and \(\tau(x)=kx+2\delta\), which can move every
answer. As an architectural intervention, we train 20 \addonly{} models in
which each language's input-number embeddings are tied to its corresponding
output-number parameters. The languages remain token-disjoint and no
parameters are tied across languages. Within each language, however, tying these embeddings forces the same relabeling to apply in both roles,  eliminating the offset and making zero task-level identifiable.

Consistent with this prediction, zero is the most frequent
cross-readout token for zero-answer equations in only \(\frac{1}{40}\)
language directions under the default architecture, but in \(\frac{35}{40}\)
directions with tied interfaces. In tied models, mean answer-zero
cross-readout is \(0.856\), versus
\(0.013\) across nonzero answers. Thus, removing a specific task-preserving freedom through the
architecture produces the predicted selective alignment without adding a
new relation.

\begin{table}[t]
\centering
\setlength{\tabcolsep}{4pt}
\renewcommand{\arraystretch}{0.96}
\begin{tabular}{@{}lcc@{}}
\toprule
 & \textbf{Answer 0} & \textbf{Nonzero Ans.} \\
\midrule

\multicolumn{3}{@{}l}{\textbf{Identity Cross-Readout}} \\
\addonly{} & $\hphantom{-}.016$\sd{.068} & $\hphantom{-}.011$\sd{.009} \\
\addcopy{} & $\hphantom{-}.055$\sd{.221} & $\hphantom{-}.012$\sd{.007} \\
\subonly{} & \focal{$\hphantom{-}.775$\sd{.248}} & $\hphantom{-}.011$\sd{.014} \\
\addsub{} & \focal{$\hphantom{-}.658$\sd{.386}} & $\hphantom{-}.006$\sd{.006} \\
\addmax{} & \focal{$\hphantom{-}.273$\sd{.370}} & \focal{$\hphantom{-}.123$\sd{.075}} \\

\addlinespace[3pt]
\multicolumn{3}{@{}l}{\textbf{Matched-Answer Cosine}} \\
\addonly{} & $\hphantom{-}.051$\sd{.405} & $\hphantom{-}.015$\sd{.073} \\
\addcopy{} & $-.083$\sd{.482} & $\hphantom{-}.003$\sd{.014} \\
\subonly{} & \focal{$\hphantom{-}.907$\sd{.047}} & $-.016$\sd{.026} \\
\addsub{} & \focal{$\hphantom{-}.792$\sd{.367}} & $\hphantom{-}.032$\sd{.126} \\
\addmax{} & \focal{$\hphantom{-}.875$\sd{.240}} & \focal{$\hphantom{-}.843$\sd{.222}} \\

\bottomrule
\end{tabular}
\caption{Answer-level cross-lingual alignment across 20 bilingual models per
condition. Cells report means \(\pm\) standard deviations across models. For each model, \emph{Nonzero Answers} averages the 112 answer-specific scores for \(c\in\{1,\ldots,112\}\); the table then reports the mean and standard deviation of these model-level averages. Identity cross-readout uses held-out
operand pairs and the other language's 113 output-number tokens.}

\label{tab:symmetry-answer-breakdown}
\end{table}

\begin{table}[t]
\centering
\setlength{\tabcolsep}{4pt}
\begin{tabular}{@{}lcc@{}}
\toprule
\textbf{Pre-Training} & \textbf{Target 0} & \textbf{Target 1} \\
\midrule
\multicolumn{3}{@{}l}{\textbf{Full Fine-Tuning}} \\
\addonly{}        & $.494$\sd{.085}         & $.492$\sd{.117} \\
\addcopy{} & $.490$\sd{.118}         & $.545$\sd{.119} \\
\subonly{}        & \focal{$.852$\sd{.065}} & $.488$\sd{.078} \\
\addsub{}         & \focal{$.808$\sd{.144}} & $.521$\sd{.102} \\
\addmax{}         & \focal{$.778$\sd{.125}} & \focal{$.772$\sd{.107}} \\
\midrule
\multicolumn{3}{@{}l}{\textbf{Frozen Body}} \\
\addonly{}        & $.502$\sd{.072}         & $.502$\sd{.103} \\
\addcopy{} & $.493$\sd{.104}         & $.518$\sd{.121} \\
\subonly{}        & \focal{$.843$\sd{.108}} & $.502$\sd{.059} \\
\addsub{}         & \focal{$.857$\sd{.172}} & $.505$\sd{.095} \\
\addmax{}         & \focal{$.759$\sd{.158}} & \focal{$.735$\sd{.146}} \\
\bottomrule
\end{tabular}
\caption{Zero-shot cross-lingual transfer for two balanced downstream
properties. Targets 0 and 1 denote $\mathrm{sum}=0/1$ for addition-based conditions and $\mathrm{diff}=0/1$ for \subonly{}. For each pre-trained model, we average five repetitions within each language direction and then average the resulting L0$\rightarrow$L1 and L1$\rightarrow$L0 target-language test accuracies. Cells report means $\pm$ standard deviations over 20 models. Chance accuracy is $0.5$.\vspace{-.5\baselineskip}}
\label{tab:zero-shot-transfer}
\end{table}

\section{Zero-Shot Cross-Lingual Transfer}
We have shown that the available relations determine which number identities
are fixed at the task level, and that learned alignment varies accordingly. We now ask whether these representational differences predict model behavior. The canonical way to measure cross-lingual generalization is through \emph{zero-shot cross-lingual transfer}: a pre-trained model is fine-tuned on a downstream task
in one language and then evaluated on the same task in another language, without receiving downstream supervision in the target language.

To study zero-shot transfer in our setting, we use the bilingual models as pre-trained base models and train a shared binary classifier using supervision from only one language. Since one classifier serves both languages but is fitted on source-language representations only, it can succeed on the target language only if the property it relies on is encoded compatibly across languages. We therefore expect poor zero-shot
transfer performance from \addonly{} and \addcopy{}, where corresponding answers remain misaligned. For \subonly{} and \addsub{}, we expect transfer to depend on the
target property: properties involving zero should transfer more readily than those involving nonzero answers. Finally, we expect broader transfer from \addmax{}, where corresponding answers are substantially more aligned across numbers.

\paragraph{Experimental Setup.}
We define two balanced downstream tasks: predicting whether the modular sum equals zero or one in the addition-based conditions, and whether the modular difference equals zero or one in \subonly{}. For each
target, we partition the 113 possible first-operand identities into disjoint
training, validation, and test sets using a \(60\%/20\%/20\%\) split. Each
first operand contributes one positive example, whose result equals the target,
and one negative example with a non-target result. This construction balances the
labels while preventing both operand-pair leakage and first-operand leakage
across splits.

For each pre-trained model, we attach a shared, randomly initialized linear
classifier to the final residual-stream activation at the equals position. We
fine-tune on source-language samples and select the
checkpoint with the lowest validation loss in that language. The target
language is not used for training or model selection and is evaluated only
after checkpoint selection. We run this process 
in both directions, using each language once as the source and once as the target language.

We consider two fine-tuning regimes. In the \emph{full} regime, both the
Transformer and classifier are updated; in the \emph{frozen-body} regime, only
the classifier is trained. For each pre-trained model and downstream property, we use five independently generated data splits and classifier initializations. Within each repetition,
the two transfer directions use the same split and initialization. We first
average repetitions within direction, then average the two directions within
each pre-trained model, and then aggregate over 20 pre-trained models per
condition.
We fine-tune for 1,000 steps and select the checkpoint with 
lowest
source-language validation loss, evaluated every 25 steps. %
See App.\ \ref{app:transfer-details}.

\paragraph{Transfer Mirrors Alignment.}
The results mirror the pattern established during
pre-training (\Cref{tab:zero-shot-transfer}). With full fine-tuning,
\addonly{} and \addcopy{} remain near chance for both downstream tasks.
With
\addsub{}, transfer becomes selective: accuracy reaches
\(0.808\) for
\(\mathrm{sum}=0\) but remains at \(0.521\) for
\(\mathrm{sum}=1\), con\-sis\-tent with the privileged alignment of zero. Yet,
\addmax{} supports broad transfer, reaching \(0.778\) for
\(\mathrm{sum}=0\) and \(0.772\) for \(\mathrm{sum}=1\).

The same qualitative pattern persists with a
frozen Transformer body:
\addsub{} transfers strongly for \(\mathrm{sum}=0\) but not
for \(\mathrm{sum}=1\), whereas \addmax{} transfers for both properties. This
shows that the transfer pattern is already present in the pre-trained
representations rather than being created only through fine-tuning.

\section{Discussion}
\label{sec:discussion}
Our results distinguish two forms of cross-lingual sharing that are often
treated as equivalent. Across all conditions, the two languages rely on closely overlapping, causally relevant answer subspaces. Yet this shared structure does not by itself align corresponding numbers: under addition alone, number-level alignment remains near chance.

Additional relational constraints can influence
which identities align. Subtraction makes zero task-level identifiable and induces selective zero alignment, whereas the order-sensitive \(\max\) produces broad alignment across numbers. The same pattern appears in zero-shot cross-lingual transfer. These results support the Relational Identifiability Hypothesis: corresponding entities are more likely to align when the available relations rule out alternative correspondences for those entities.

Relational structure is not the only source of alignment. In natural multi\-lin\-gu\-al data, shared tokens, parallel passages,
code-\-swit\-ch\-ing, and correlated contexts provide more direct evidence about how meanings correspond. 
Hence, we do not claim that 
relational constraints replace lexical or parallel anchors, but that they constitute an \emph{independent source of alignment}: even without direct cross-lingual evidence, relations can rule out alternative correspondences for particular entities.

Moreover, we 
predict that \textit{uneven alignment} can occur across 
entities and languages. Alignment can also 
depend on whether the 
training data provides enough evidence for 
models to recover 
relevant relations 
that encourage
cross-lingual correspondences. We therefore expect sparsely observed entities and lower-resourced languages %
to align less reliably.

Our mathematical world offers a controlled instance of a more general
principle. Natural languages describe worlds composed not only of entities,
but also of properties and relations that connect them. In real
MLLMs, these structures are entangled with many other alignment sources
that cannot be easily manipulated independently. The arithmetic
setting lets us isolate relational constraints and observe their
representational consequences directly. Moreover, 
related work has identified
comparable arithmetic mechanisms in pre-trained language models \citep{zhou2024pretrained,kantamneni2025language}, showing that the
representations we study in this work %
are not limited to these toy
models. Our experiments suggest that a shared computational space can emerge before a model resolves which entities correspond across languages. Learning more about how entities relate and are grounded can help resolve this ambiguity and promote alignment.

\section{Conclusion}
Inspired by the Platonic Representation Hypothesis, we ask whether two
languages with disjoint vocabularies, learned jointly by a shared
Transformer, can develop aligned representations of corresponding 
entities. We propose the \emph{Relational Identifiability Hypothesis}: relations in a
shared world can promote cross-lingual entity alignment by ruling out
alternative correspondences.

Our bilingual arithmetic world reveals a two-faceted 
picture. Across all conditions, the languages develop strongly overlapping, causally relevant answer structure. Yet under modular addition alone, corresponding numbers remain unaligned: shared parameters, successful generalization, and shared computational structure are insufficient for number-level identity alignment. Additional relational constraints can make particular identities task-level identifiable and change learned alignment accordingly: subtraction fixes zero and induces selective zero alignment, while an order-sensitive operation produces broad alignment across numbers.
Despite their synthetic scope, these results help clarify the multilingual debate over lexical anchors. Shared tokens and parallel data provide direct evidence of cross-lingual correspondence, but they are not the only possible source of alignment. In our controlled setting, relations alone can rule out alternative correspondences between token-disjoint languages.

\clearpage

\bibliography{tacl2021,tacl/anthology-1,tacl/anthology-2}

\begin{thebibliography}{24}
\expandafter\ifx\csname natexlab\endcsname\relax\def\natexlab#1{#1}\fi

\bibitem[{Cao et~al.(2020)Cao, Kitaev, and Klein}]{Cao2020Multilingual}
Steven Cao, Nikita Kitaev, and Dan Klein. 2020.
\newblock \href {https://openreview.net/forum?id=r1xCMyBtPS} {Multilingual alignment of contextual word representations}.
\newblock In \emph{International Conference on Learning Representations}.

\bibitem[{Conneau et~al.(2020)Conneau, Wu, Li, Zettlemoyer, and Stoyanov}]{conneau-etal-2020-emerging}
Alexis Conneau, Shijie Wu, Haoran Li, Luke Zettlemoyer, and Veselin Stoyanov. 2020.
\newblock \href {https://doi.org/10.18653/v1/2020.acl-main.536} {Emerging cross-lingual structure in pretrained language models}.
\newblock In \emph{Proceedings of the 58th Annual Meeting of the Association for Computational Linguistics}, pages 6022--6034, Online. Association for Computational Linguistics.

\bibitem[{Deshpande et~al.(2022)Deshpande, Talukdar, and Narasimhan}]{deshpande-etal-2022-bert}
Ameet Deshpande, Partha Talukdar, and Karthik Narasimhan. 2022.
\newblock \href {https://doi.org/10.18653/v1/2022.naacl-main.264} {When is {BERT} multilingual? isolating crucial ingredients for cross-lingual transfer}.
\newblock In \emph{Proceedings of the 2022 Conference of the North American Chapter of the Association for Computational Linguistics: Human Language Technologies}, pages 3610--3623, Seattle, United States. Association for Computational Linguistics.

\bibitem[{Dufter and Sch{\"u}tze(2020)}]{dufter-schutze-2020-identifying}
Philipp Dufter and Hinrich Sch{\"u}tze. 2020.
\newblock \href {https://doi.org/10.18653/v1/2020.emnlp-main.358} {Identifying elements essential for {BERT}{'}s multilinguality}.
\newblock In \emph{Proceedings of the 2020 Conference on Empirical Methods in Natural Language Processing (EMNLP)}, pages 4423--4437, Online. Association for Computational Linguistics.

\bibitem[{Dumas et~al.(2025)Dumas, Wendler, Veselovsky, Monea, and West}]{dumas-etal-2025-separating}
Cl{\'e}ment Dumas, Chris Wendler, Veniamin Veselovsky, Giovanni Monea, and Robert West. 2025.
\newblock \href {https://doi.org/10.18653/v1/2025.acl-long.1536} {Separating tongue from thought: Activation patching reveals language-agnostic concept representations in transformers}.
\newblock In \emph{Proceedings of the 63rd Annual Meeting of the Association for Computational Linguistics (Volume 1: Long Papers)}, pages 31822--31841, Vienna, Austria. Association for Computational Linguistics.

\bibitem[{Huh et~al.(2024)Huh, Cheung, Wang, and Isola}]{pmlr-v235-huh24a}
Minyoung Huh, Brian Cheung, Tongzhou Wang, and Phillip Isola. 2024.
\newblock \href {https://proceedings.mlr.press/v235/huh24a.html} {Position: The platonic representation hypothesis}.
\newblock In \emph{Proceedings of the 41st International Conference on Machine Learning}, volume 235 of \emph{Proceedings of Machine Learning Research}, pages 20617--20642. PMLR.

\bibitem[{Jha et~al.(2025)Jha, Zhang, Shmatikov, and Morris}]{jha2025harnessing}
Rishi~Dev Jha, Collin Zhang, Vitaly Shmatikov, and John~Xavier Morris. 2025.
\newblock \href {https://openreview.net/forum?id=jiCLUPq5xv} {Harnessing the universal geometry of embeddings}.
\newblock In \emph{The Thirty-ninth Annual Conference on Neural Information Processing Systems}.

\bibitem[{K et~al.(2020)K, Wang, Mayhew, and Roth}]{ketal}
Karthikeyan K, Zihan Wang, Stephen Mayhew, and Dan Roth. 2020.
\newblock \href {https://openreview.net/forum?id=HJeT3yrtDr} {Cross-lingual ability of multilingual {BERT:} an empirical study}.
\newblock In \emph{8th International Conference on Learning Representations, {ICLR} 2020, Addis Ababa, Ethiopia, April 26-30, 2020}. OpenReview.net.

\bibitem[{Kallini et~al.(2025)Kallini, Jurafsky, Potts, and Bartelds}]{kallini-etal-2025-false}
Julie Kallini, Dan Jurafsky, Christopher Potts, and Martijn Bartelds. 2025.
\newblock \href {https://doi.org/10.18653/v1/2025.findings-emnlp.1153} {False {F}riends are not foes: Investigating vocabulary overlap in multilingual language models}.
\newblock In \emph{Findings of the Association for Computational Linguistics: EMNLP 2025}, pages 21138--21154, Suzhou, China. Association for Computational Linguistics.

\bibitem[{Kantamneni and Tegmark(2025)}]{kantamneni2025language}
Subhash Kantamneni and Max Tegmark. 2025.
\newblock \href {https://openreview.net/forum?id=CqViN4dQJk} {Language models use trigonometry to do addition}.
\newblock In \emph{ICLR 2025 Workshop on Building Trust in Language Models and Applications}.

\bibitem[{Lauscher et~al.(2020)Lauscher, Ravishankar, Vuli{\'c}, and Glava{\v{s}}}]{lauscher-etal-2020-zero}
Anne Lauscher, Vinit Ravishankar, Ivan Vuli{\'c}, and Goran Glava{\v{s}}. 2020.
\newblock \href {https://doi.org/10.18653/v1/2020.emnlp-main.363} {From zero to hero: {O}n the limitations of zero-shot language transfer with multilingual {T}ransformers}.
\newblock In \emph{Proceedings of the 2020 Conference on Empirical Methods in Natural Language Processing (EMNLP)}, pages 4483--4499, Online. Association for Computational Linguistics.

\bibitem[{Liang et~al.(2023)Liang, Gonen, Mao, Hou, Goyal, Ghazvininejad, Zettlemoyer, and Khabsa}]{liang-etal-2023-xlm}
Davis Liang, Hila Gonen, Yuning Mao, Rui Hou, Naman Goyal, Marjan Ghazvininejad, Luke Zettlemoyer, and Madian Khabsa. 2023.
\newblock \href {https://doi.org/10.18653/v1/2023.emnlp-main.813} {{XLM}-{V}: Overcoming the vocabulary bottleneck in multilingual masked language models}.
\newblock In \emph{Proceedings of the 2023 Conference on Empirical Methods in Natural Language Processing}, pages 13142--13152, Singapore. Association for Computational Linguistics.

\bibitem[{Lin et~al.(2019)Lin, Chen, Lee, Li, Zhang, Xia, Rijhwani, He, Zhang, Ma, Anastasopoulos, Littell, and Neubig}]{lin-etal-2019-choosing}
Yu-Hsiang Lin, Chian-Yu Chen, Jean Lee, Zirui Li, Yuyan Zhang, Mengzhou Xia, Shruti Rijhwani, Junxian He, Zhisong Zhang, Xuezhe Ma, Antonios Anastasopoulos, Patrick Littell, and Graham Neubig. 2019.
\newblock \href {https://doi.org/10.18653/v1/P19-1301} {Choosing transfer languages for cross-lingual learning}.
\newblock In \emph{Proceedings of the 57th Annual Meeting of the Association for Computational Linguistics}, pages 3125--3135, Florence, Italy. Association for Computational Linguistics.

\bibitem[{Nanda et~al.(2023)Nanda, Chan, Lieberum, Smith, and Steinhardt}]{nanda2023progress}
Neel Nanda, Lawrence Chan, Tom Lieberum, Jess Smith, and Jacob Steinhardt. 2023.
\newblock \href {https://openreview.net/forum?id=9XFSbDPmdW} {Progress measures for grokking via mechanistic interpretability}.
\newblock In \emph{The Eleventh International Conference on Learning Representations}.

\bibitem[{Parcalabescu et~al.(2021)Parcalabescu, Trost, and Frank}]{parcalabescu-etal-2021-multimodality}
Letitia Parcalabescu, Nils Trost, and Anette Frank. 2021.
\newblock \href {https://aclanthology.org/2021.mmsr-1.1/} {What is multimodality?}
\newblock In \emph{Proceedings of the 1st Workshop on Multimodal Semantic Representations (MMSR)}, pages 1--10, Groningen, Netherlands (Online). Association for Computational Linguistics.

\bibitem[{Pires et~al.(2019)Pires, Schlinger, and Garrette}]{pires-etal-2019-multilingual}
Telmo Pires, Eva Schlinger, and Dan Garrette. 2019.
\newblock \href {https://doi.org/10.18653/v1/P19-1493} {How multilingual is multilingual {BERT}?}
\newblock In \emph{Proceedings of the 57th Annual Meeting of the Association for Computational Linguistics}, pages 4996--5001, Florence, Italy. Association for Computational Linguistics.

\bibitem[{Riemenschneider and Frank(2025)}]{riemenschneider-frank-2025-cross}
Frederick Riemenschneider and Anette Frank. 2025.
\newblock \href {https://doi.org/10.18653/v1/2025.acl-long.661} {Cross-lingual generalization and compression: From language-specific to shared neurons}.
\newblock In \emph{Proceedings of the 63rd Annual Meeting of the Association for Computational Linguistics (Volume 1: Long Papers)}, pages 13470--13491, Vienna, Austria. Association for Computational Linguistics.

\bibitem[{Sch{\"{a}}fer et~al.(2024)Sch{\"{a}}fer, Ravfogel, Hofmann, Pimentel, and Schlag}]{schaeferetal}
Anton Sch{\"{a}}fer, Shauli Ravfogel, Thomas Hofmann, Tiago Pimentel, and Imanol Schlag. 2024.
\newblock \href {https://doi.org/10.48550/ARXIV.2404.07982} {Language imbalance can boost cross-lingual generalisation}.
\newblock \emph{CoRR}, abs/2404.07982v1.

\bibitem[{Touvron et~al.(2023)Touvron, Martin, Stone, Albert, Almahairi, Babaei, Bashlykov, Batra, Bhargava, Bhosale et~al.}]{touvron2023llama}
Hugo Touvron, Louis Martin, Kevin Stone, Peter Albert, Amjad Almahairi, Yasmine Babaei, Nikolay Bashlykov, Soumya Batra, Prajjwal Bhargava, Shruti Bhosale, et~al. 2023.
\newblock Llama 2: Open foundation and fine-tuned chat models.
\newblock \emph{arXiv preprint arXiv:2307.09288v1}.

\bibitem[{Wendler et~al.(2024)Wendler, Veselovsky, Monea, and West}]{wendler-etal-2024-llamas}
Chris Wendler, Veniamin Veselovsky, Giovanni Monea, and Robert West. 2024.
\newblock \href {https://doi.org/10.18653/v1/2024.acl-long.820} {Do llamas work in {E}nglish? on the latent language of multilingual transformers}.
\newblock In \emph{Proceedings of the 62nd Annual Meeting of the Association for Computational Linguistics (Volume 1: Long Papers)}, pages 15366--15394, Bangkok, Thailand. Association for Computational Linguistics.

\bibitem[{Wu and Dredze(2019)}]{wu-dredze-2019-beto}
Shijie Wu and Mark Dredze. 2019.
\newblock \href {https://doi.org/10.18653/v1/D19-1077} {Beto, bentz, becas: The surprising cross-lingual effectiveness of {BERT}}.
\newblock In \emph{Proceedings of the 2019 Conference on Empirical Methods in Natural Language Processing and the 9th International Joint Conference on Natural Language Processing (EMNLP-IJCNLP)}, pages 833--844, Hong Kong, China. Association for Computational Linguistics.

\bibitem[{Zhong et~al.(2023)Zhong, Liu, Tegmark, and Andreas}]{zhong2023the}
Ziqian Zhong, Ziming Liu, Max Tegmark, and Jacob Andreas. 2023.
\newblock \href {https://openreview.net/forum?id=S5wmbQc1We} {The clock and the pizza: Two stories in mechanistic explanation of neural networks}.
\newblock In \emph{Thirty-seventh Conference on Neural Information Processing Systems}.

\bibitem[{Zhou et~al.(2024)Zhou, Fu, Sharan, and Jia}]{zhou2024pretrained}
Tianyi Zhou, Deqing Fu, Vatsal Sharan, and Robin Jia. 2024.
\newblock \href {https://openreview.net/forum?id=i4MutM2TZb} {Pre-trained large language models use fourier features to compute addition}.
\newblock In \emph{The Thirty-eighth Annual Conference on Neural Information Processing Systems}.

\bibitem[{Ziyin and Chuang(2025)}]{ziyin2025proof}
Liu Ziyin and Isaac Chuang. 2025.
\newblock Proof of a perfect platonic representation hypothesis.
\newblock \emph{arXiv preprint arXiv:2507.01098v1}.

\end{thebibliography}
\bibliographystyle{acl_natbib}

\onecolumn

\appendix

\section{Causal Fourier Validation}
\label{app:fourier-validation}
We verify that the successfully trained models use a compact Fourier structure
to solve their modular arithmetic tasks. This analysis is performed separately
for each language and task.

For each language and task, we collect the MLP activation at the equals
position for all \(p^2\) operand pairs and arrange these activations in a
\(p\times p\) grid indexed by \(a\) and \(b\). A two-dimensional Fourier
transform decomposes this grid into repeating patterns over the operands. For
addition, we examine patterns that vary with the modular sum \(a+b\); for
subtraction, we examine patterns that vary with the modular difference \(a-b\).
We select the five frequencies that carry the highest MLP-activation 
in the
relevant pattern. These frequencies describe global patterns across all
numbers; they are not assigned to individual number identities.

Observing these components does not by itself establish that the model uses
them. We therefore
test the selected components through two causal interventions.

In the \emph{retention} intervention, we discard all variation in the MLP
activation except its mean and the five selected Fourier components. We then
ask whether the model can still solve the task using only this clean
reconstruction. In the \emph{removal} intervention, we do the reverse: we
preserve the original activation except for the five selected components and
ask whether performance collapses. Both interventions are evaluated over the
complete \(p^2\) operand grid.

As shown in \Cref{tab:appendix-fourier-validation}, the retained
reconstructions preserve nearly all performance, with mean accuracy between
\(0.941\) and \(1.000\) across conditions and tasks. Conversely, removing the
same components produces mean accuracy drops between \(0.987\) and \(0.996\).
For every analyzed random seed, the accuracy drop, averaged across its two
languages, is at least \(0.961\). Thus, a small set of Fourier components is
both sufficient to preserve the model's solution and necessary for its
performance.

This analysis validates a causal Fourier structure for each language in our synthetic multilingual construction of modular addition and subtraction.
It does not establish that the two languages select the same
frequencies or implement identical complete circuits. We also do not expect the same sum- or difference-based Fourier components for 
\(\max(a,b)\), because its answer depends on numerical ordering rather
than on a modular sum or difference. In the \addmax{} condition, we therefore
apply the Fourier validation only to addition.

\section{Model and Training Details}
\label{app:training-details}

\subsection{Architecture, Vocabulary, and Data}
We use a one-layer decoder-only Transformer with the architecture and training
hyperparameters listed in \Cref{tab:appendix-default-config}. Each input sequence consists of a task token, two number tokens, and an equals
token. An addition input in language \(\ell\) has the form
\enquote{\(
\textsc{[modular\_add]}_{\ell}\;
a_{\ell}\;
b_{\ell}\;
=_{\ell}.
\)
}
For example,
\enquote{\(
\textsc{[modular\_add]}_{\ell}\;
1_{\ell}\;
2_{\ell}\;
=_{\ell}
\)
}
asks the model to predict the language-\(\ell\) output token representing
\(3\). The two languages use separate task, number, equals, and output tokens. Within
each language, the same number vocabulary is shared across that language's
active tasks. The Transformer parameters are shared across both languages and
all tasks.

For each condition, we randomly select \(30\%\) of the \(p^2\) input pairs for training and reserve the remaining \(70\%\) for evaluation. The same
split is used across both languages and, in multitask conditions, across all
active tasks. Thus, differences between languages or tasks cannot be attributed
to different input-pair distributions.

\begin{table}[h]
\centering
\small
\setlength{\tabcolsep}{4pt}
\begin{tabular}{@{}lr@{}}
\toprule
\textbf{Parameter} & \textbf{Value} \\
\midrule
Prime modulus \(p\) & 113 \\
Transformer layers & 1 \\
Attention heads & 4 \\
\(d_{\mathrm{model}}\) & 128 \\
\(d_{\mathrm{head}}\) & 32 \\
\(d_{\mathrm{MLP}}\) & 512 \\
Activation & ReLU \\
Training fraction & \(0.3\) \\
Optimizer & AdamW \\
Learning rate & \(10^{-3}\) \\
Weight decay & \(1.0\) \\
Adam \((\beta_1,\beta_2)\) & \((0.9,0.98)\) \\
Training steps & 50,000 \\
\bottomrule
\end{tabular}
\caption{Architecture and training hyperparameters used across conditions. Each update processes the complete training split for every active task-language combination.}
\label{tab:appendix-default-config}
\end{table}

\subsection{Experimental Conditions}
We train models under five conditions that vary the relations expressed by the
training tasks.
In \addonly{}, both languages are trained on
modular addition. In \addcopy{}, each language is trained on two copies of
modular addition that use distinct task tokens but implement the same relation.
This condition controls for the additional examples and optimization signal
introduced by a second task without adding new relational structure.

In \addsub{}, each language is trained jointly on modular addition and modular
subtraction. The \subonly{} condition contains modular subtraction alone and lets us test
whether the effects observed in \addsub{} arise from subtraction itself rather
than from the interaction between addition and subtraction.
 Finally, in
\addmax{}, each language is trained jointly on modular addition and the
ordinary maximum relation. Each optimization step processes the complete training split for every active task-language combination.

Across conditions, the architecture, vocabulary design, operand-pair split, and
optimization hyperparameters remain fixed. The conditions therefore isolate
the effect of changing which relations are available during pre-training.

\subsection{Model Selection and Training Stability}
\label{app:grokking-details}

We analyze the final checkpoint after 50,000 training steps. A model is eligible
for analysis if it reaches at least \(98\%\) held-out accuracy in every active
task-language combination. For each condition, we fix the first 20 eligible models before conducting our analyses.

The five conditions differ substantially in training stability. Achieving
grokking is not equally reliable across runs, and the difficulty depends on the training setup, including both the number of languages and the task being learned. Conditions with multiple tasks tend to require more runs, but task identity also matters: \subonly{} is already less stable than \addonly{}, and \addmax{} requires the largest number of runs before 20 models meet the eligibility criterion. \Cref{tab:appendix-training-stability} reports the number of runs required in each condition.

Excluded runs exhibit different failure modes across conditions. Most excluded \addmax{} runs learned all tasks but fell slightly below our eligibility threshold, reaching a median minimum held-out accuracy of \(97.4\%\). By contrast, all excluded \subonly{} runs failed to grok in at least one language. Our \(98\%\) eligibility threshold is intentionally strict because we aim to compare models that have mastered all active task-language combinations, rather than risk attributing differences in our mechanistic analyses to differences in task performance. We apply this threshold uniformly and stop once the 20th eligible model is obtained.

\begin{table}[t]
\centering
\small
\setlength{\tabcolsep}{4pt}
\begin{tabular}{@{}lrr@{}}
\toprule
\textbf{Condition} & \textbf{Runs to 20 eligible models} & \textbf{Failed runs} \\
\midrule
\addonly{} & 20 & 0 \\
\addcopy{} & 20 & 0 \\
\subonly{} & 51 & 31 \\
\addsub{} & 21 & 1 \\
\addmax{} & 72 & 52 \\
\bottomrule
\end{tabular}
\caption{Condition-specific training stability. Runs are processed in seed order, and counting stops once the 20th eligible model is obtained. Eligibility requires at least $98\%$ held-out accuracy in every active task--language combination.}
\label{tab:appendix-training-stability}
\end{table}

\section{Cross-Lingual Alignment Analyses}
\label{app:alignment-details}

\subsection{Cross-Readout and Cosine Similarity}
We use two complementary diagnostics to test whether corresponding numbers align across languages. Cross-readout tests whether one language's output representations can be read using the other language's output vocabulary, whereas cosine similarity compares the residual-stream directions associated with corresponding answers.

\paragraph{Cross-Readout.}
For each language direction \(\ell_s \rightarrow \ell_t\), we evaluate held-out inputs expressed in the source language \(\ell_s\) and restrict the resulting logits to the \(p\) output-number tokens of the target language \(\ell_t\). Cross-readout is correct when the highest-logit target-language token denotes the ground-truth answer to the source-language input. Because the candidate set contains \(p=113\) tokens, chance accuracy is \(\frac{1}{113}\). We compute accuracy separately for each ground-truth answer \(c\). The broad cross-readout score is the unweighted mean of these 113 answer-specific accuracies, followed by an average over the two language directions. This metric therefore tests whether a hidden state produced for one language can be read directly using the other language's unembedding directions.

\paragraph{Cosine Similarity.}
For each language, we record the final residual-stream activation at the equals
position for every input pair. We group these activations by their ground-truth
answer and average within each group, yielding one answer signature
\(\mathbf{s}_{\ell}(c)\) for every language \(\ell\) and answer
\(c\in\{0,\ldots,p-1\}\). We center these signatures by subtracting the mean
signature within each language, thereby removing language-wide offsets.

We then compute the average cosine similarity between the centered signatures
of corresponding answers:
\[
\frac{1}{p}\sum_{c=0}^{p-1}
\cos\!\left(\tilde{\mathbf{s}}_{\ell_0}(c),
            \tilde{\mathbf{s}}_{\ell_1}(c)\right),
\]
where \(\tilde{\mathbf{s}}_{\ell}(c)\) denotes the centered signature.  High
cosine similarity indicates that corresponding numerical answers occupy
similar directions in the residual stream. Unlike cross-readout, this metric does not depend on either language's unembedding directions.

\subsection{Answer-Subspace Estimation}

Matched-answer cosine requires corresponding answers to point in similar
directions. The two languages could instead represent individual answers
differently while encoding answer information within the same low-dimensional
subspace. We therefore estimate an answer subspace separately for each
language.

For language \(\ell\), we stack its centered answer signatures into a matrix
\(
\tilde{\mathbf{S}}_{\ell}
\in \mathbb{R}^{p\times d_{\mathrm{model}}},
\)
with one row for each answer \(c\). We apply singular value decomposition,
\(
\tilde{\mathbf{S}}_{\ell}
=
\mathbf{U}_{\ell}\mathbf{\Sigma}_{\ell}\mathbf{V}_{\ell}^{\top},
\)
and define the rank-\(r\) answer subspace as the span of the first \(r=8\)
right-singular vectors in \(\mathbf{V}_{\ell}\). 

To compare the two languages, let
\(\mathbf{B}_{\ell_0},\mathbf{B}_{\ell_1}
\in\mathbb{R}^{d_{\mathrm{model}}\times r}\)
denote orthonormal bases for their answer subspaces. The singular values of
\(\mathbf{B}_{\ell_0}^{\top}\mathbf{B}_{\ell_1}\) are the cosines of the
canonical angles between the subspaces. We define subspace overlap as their
mean:
\[
\operatorname{overlap}(\ell_0,\ell_1)
=
\frac{1}{r}\sum_{i=1}^{r}
\sigma_i\!\left(
\mathbf{B}_{\ell_0}^{\top}\mathbf{B}_{\ell_1}
\right).
\]
An overlap of \(1\) indicates identical subspaces, whereas smaller values
indicate that the answer signatures vary along different residual-stream
directions. Because two low-dimensional subspaces can overlap partly by
chance, we compare the observed values with isotropically sampled random
subspaces of the same rank and with subspaces obtained from independently
trained models.

\subsection{Subspace Ablation and Controls}
\label{app:subspace-ablation}

Subspace overlap alone does not establish that the identified directions are
used by the model. To analyze the model's internal behavior, we 
intervene on the residual stream by removing
its projection onto an estimated answer subspace. For these interventions, we
estimate the answer signatures and their SVD basis using training samples
only, and measure the resulting accuracy change on held-out data samples.

Let \(\ell\) denote the language being evaluated, \(\ell'\) the other
language, and \(s\) the intervention site. For each language and site, we
estimate a mean training activation \(\boldsymbol{\mu}_{\ell,s}\) and an
orthonormal rank-\(8\) answer-subspace basis
\(\mathbf{B}_{\ell,s}\). In the other-language intervention, we replace the
equals-position activation \(\mathbf{x}\) with
\[
\mathbf{x}'
=
\mathbf{x}
-
\mathbf{B}_{\ell',s}^{}\mathbf{B}_{\ell',s}^{\top}
\bigl(\mathbf{x}-\boldsymbol{\mu}_{\ell,s}^{}\bigr).
\]
Thus, the intervention removes variation along the selected directions while
preserving the language's mean activation. We apply the intervention at two
sites: the residual stream immediately before the MLP and the final residual
stream after the MLP. Accuracy drop is defined as the held-out accuracy before the
intervention minus the held-out accuracy after the intervention.

We consider four choices of \(\mathbf{B}\). The \emph{own-language} condition
uses the answer subspace estimated from the evaluated language.
The \emph{other-language} condition instead uses the subspace estimated from
the other language in the same bilingual model. A large drop in the \textit{other-language} 
condition
shows that directions identified from one language are functionally important
for the other. The \emph{independent-model} control uses the corresponding
language's subspace from a separately initialized model, testing whether any
model that solves the task develops causally interchangeable directions.
Finally, the \emph{random-subspace} control removes an isotropically sampled
subspace of the same rank.

For each model, we average over both evaluated languages and, for the random
control, over five independently sampled subspaces. \Cref{tab:appendix-subspace-ablation}
reports the mean and standard deviation of the resulting model-level accuracy
drops across the 20 seeds.

\subsection{Answer-Level Alignment}

The aggregate cross-readout and cosine scores average over all \(p=113\)
answers and can therefore conceal alignment confined to particular values. We
additionally compute both metrics separately for each answer \(c\). For
cross-readout, we restrict the held-out examples to those whose ground-truth
answer is \(c\) and average accuracy over the two language directions. For
cosine similarity, we use the cosine between the two languages' centered
signatures for answer \(c\).

Because our hypotheses assign a special role to zero, we report answer \(0\)
separately. For each model, the nonzero score is the unweighted mean of the
112 answer-specific scores for \(c\in\{1,\ldots,112\}\). Overall means and
standard deviations are then computed over the 20 model-level values.

\subsection{Direction- %
and Model-Specific Results}

Averaging the two cross-readout directions can conceal asymmetric
correspondences. We therefore examine answer-zero cross-readout separately in
each direction for the \subonly{} models. For each direction, we consider all
\(p=113\) equations of the form \(a-a=0\) and identify the target-language
token predicted most frequently across these equations. We say that 
\enquote{zero maps to zero} when this most frequent prediction is the target-language token for zero.

Of the 20 models, 11 map zero to zero in both directions and nine do so in
exactly one direction. No model fails to map zero to zero correctly in both
directions. The asymmetry is not explained by which language groks first.
Among the nine asymmetric models, the correctly aligned direction uses the
earlier-grokking language as its source in five cases and as its target in
four.

The incorrect directions nevertheless implement stable correspondences rather
than producing unrelated predictions across equations. For a given direction,
we define \emph{mapping consistency} as the fraction of the \(p\) equations
\(a-a=0\) assigned to that direction's most frequently predicted
target-language token. Across the nine incorrect directions, mapping
consistency is \(0.951\pm0.086\).
Thus, when identity cross-readout fails, an individual model generally maps
zero consistently to one model-specific alternative token.

\section{Zero-Shot Cross-Lingual Transfer}
\label{app:transfer-details}

\subsection{Downstream Tasks and Data Splits}

We construct binary classification tasks from the arithmetic sequences used
during pre-training. In Task 1, for models that contain the addition condition, %
the classifier must
predict whether the modular sum equals \(0\) or \(1\). 
In Task 2, for models that use \subonly{}, the classifier must
instead predict whether the modular difference equals \(0\) or \(1\).

The classifiers receive the final residual-stream activation at the equals
position of an ordinary arithmetic input. An addition input in language
\(\ell\), for example, has the form
\enquote{\(
\textsc{[modular\_add]}_{\ell}\;
a_{\ell}\;
b_{\ell}\;
=_{\ell}.
\)
}
With %
modulo
\(113\), the sequence
\enquote{\(
\textsc{[modular\_add]}_{\ell}\;
5_{\ell}\;
108_{\ell}\;
=_{\ell}
\)
}
is a positive example for the property \(\mathrm{sum}=0\), because
\(5+108=0\pmod{113}\). The model's arithmetic output token is not included in
the classifier input; the classifier must recover the relevant property from
the residual-stream activation at the equals position.

For each target value \(t\in\{0,1\}\), we partition the \(p=113\) possible
first operands into training, validation, and test sets using a
\(60\%/20\%/20\%\) split. This gives 68 training, 23 validation, and 22 test
first operand cases. For every first operand, we construct one equation whose answer
equals \(t\) and one whose answer does not. Each split is therefore balanced,
containing 136 training, 46 validation, and 44 test instances. Because the
first-operand sets are disjoint, neither complete operand pairs nor
first-operand identities are shared across the three splits.

We generate five independent data partitions for each target value. Within
each repetition, both transfer directions use the same underlying operand
pairs, expressed using the respective language's disjoint tokens. 

\subsection{Fine-Tuning and Model Selection}

For each downstream task, we attach a linear binary classifier to the
pre-trained model's final residual-stream activation at the equals position.
The classifier maps the \(128\)-dimensional residual-stream activation to two
logits. 

We perform source-language fine-tuning separately in both directions. In an
\(\ell_0\rightarrow\ell_1\) run, the classifier and, where applicable, the
Transformer are trained using only language-\(\ell_0\) examples. After model
selection, the resulting classifier is applied unchanged to language-\(\ell_1\)
test examples. The reverse direction starts again from the original
pre-trained checkpoint and trains only on language \(\ell_1\). Within each
repetition, the two directions use the same underlying data split and the same
classifier initialization.

We consider two update regimes. Under \emph{full fine-tuning}, we update both
the pre-trained Transformer and the classifier. Under \emph{frozen-body
fine-tuning}, all Transformer parameters remain fixed and only the classifier
is updated. The latter condition tests whether the downstream property is
already accessible in compatible directions across the two languages before
fine-tuning changes the Transformer.

Each run uses full-batch AdamW optimization for 1,000 update steps, with a
learning rate of \(10^{-4}\), no weight decay, and
\((\beta_1,\beta_2)=(0.9,0.98)\). We evaluate the source-language validation
loss every 25 steps, including before the first update, and retain the
checkpoint with the lowest source-language validation loss. Training always
continues for the full 1,000 steps before the selected checkpoint is restored.

The target language is not used for gradient updates, checkpoint selection, or
any other model-selection decision. We evaluate its test accuracy only after
restoring the source-selected checkpoint. For every pre-trained model, target
property, transfer direction, and update regime, we repeat this procedure with
five independently generated data splits and classifier initializations.

\clearpage

\section{Complementary Results}

\begin{table*}[h]
\centering
\small
\setlength{\tabcolsep}{4pt}
\begin{tabular}{@{}llccc@{}}
\toprule
\textbf{Condition} & \textbf{Task} & \textbf{Full Accuracy} & \textbf{Retained Accuracy} & \textbf{Removal Drop} \\
\midrule
\addonly{} & Add & $1.000 \pm 0.000$ & $1.000 \pm 0.000$ & $0.992 \pm 0.004$ \\
\addcopy{} & Add 1 & $1.000 \pm 0.000$ & $1.000 \pm 0.000$ & $0.994 \pm 0.002$ \\
\addcopy{} & Add 2 & $1.000 \pm 0.000$ & $1.000 \pm 0.000$ & $0.994 \pm 0.002$ \\
\subonly{} & Sub & $1.000 \pm 0.000$ & $1.000 \pm 0.000$ & $0.996 \pm 0.003$ \\
\addsub{} & Add & $0.999 \pm 0.002$ & $0.992 \pm 0.023$ & $0.993 \pm 0.007$ \\
\addsub{} & Sub & $0.999 \pm 0.002$ & $0.992 \pm 0.023$ & $0.993 \pm 0.006$ \\
\addmax{} & Add & $0.995 \pm 0.003$ & $0.941 \pm 0.097$ & $0.987 \pm 0.007$ \\
\bottomrule
\end{tabular}
\caption{Causal Fourier validation for modular addition and subtraction tasks (mean $\pm$ standard deviation across 20 models; languages averaged within model). For each language and task, we select the five task-relevant frequencies carrying the most MLP-activation energy and either retain or remove their Fourier components. The \(\max\) task in \addmax{} is exempt from retention/removal as
the 
Fourier-mechanism prediction is not defined for ordinary numerical ordering.}
\label{tab:appendix-fourier-validation}
\end{table*}

\begin{table*}[h]
\centering
\small
\setlength{\tabcolsep}{5pt}
\begin{tabular}{@{}lccc@{}}
\toprule
\textbf{Condition} & \textbf{Same Bilingual Model} & \textbf{Independent Bilingual Models} & \textbf{Random Subspaces} \\
\midrule
\addonly{} & $.944 \pm 0.080$ & $0.216 \pm 0.014$ & $0.211 \pm 0.020$ \\
\addcopy{} & $.928 \pm 0.091$ & $0.210 \pm 0.019$ & $0.210 \pm 0.019$ \\
\subonly{} & $.889 \pm 0.086$ & $0.215 \pm 0.008$ & $0.210 \pm 0.019$ \\
\addsub{} & $.855 \pm 0.176$ & $0.224 \pm 0.020$ & $0.210 \pm 0.019$ \\
\addmax{} & $.924 \pm 0.086$ & $0.208 \pm 0.010$ & $0.210 \pm 0.019$ \\
\bottomrule
\end{tabular}
\caption{Rank-8 answer-subspace overlap at the final residual stream (mean $\pm$ standard deviation). Same-model values use the 20 frozen bilingual models in each condition. Independent-model controls use ten disjoint pairs of bilingual models; the two same-language comparisons are averaged within each pair. Random controls use 2,000 pairs of isotropically sampled rank-matched subspaces in $\mathbb{R}^{d_{\mathrm{model}}}$. All learned bases are estimated from arithmetic training examples only.}
\label{tab:appendix-subspace-overlap-controls}
\end{table*}

\begin{table*}[h]
\centering
\small
\setlength{\tabcolsep}{4pt}
\resizebox{\linewidth}{!}{
\begin{tabular}{@{}lccccc@{}}
\toprule
\textbf{Accuracy Drop} & \textbf{\addonly{}} & \textbf{\addcopy{}} & \textbf{\subonly{}} & \textbf{\addsub{}} &  \textbf{\addmax{}} \\
\midrule
\multicolumn{6}{@{}l}{\textbf{Pre-MLP Residual Stream}} \\
\quad Own-Language Subspace & $0.958 \pm 0.014$ & $0.956 \pm 0.026$ & $0.985 \pm 0.021$ & $0.969 \pm 0.020$ &  $0.984 \pm 0.004$ \\
\quad Other-Language Subspace & $0.928 \pm 0.036$ & $0.923 \pm 0.046$ & $0.956 \pm 0.046$ & $0.947 \pm 0.047$ &  $0.983 \pm 0.005$ \\
\quad Independent-Model Subspace & $0.001 \pm 0.001$ & $0.001 \pm 0.003$ & $0.001 \pm 0.002$ & $0.007 \pm 0.011$ &  $0.114 \pm 0.122$ \\
\quad Random Subspace & $0.001 \pm 0.001$ & $0.001 \pm 0.001$ & $0.001 \pm 0.002$ & $0.005 \pm 0.009$ &  $0.100 \pm 0.101$ \\

\multicolumn{6}{@{}l}{\textbf{Final Residual Stream}} \\
\quad Own-Language Subspace & $0.992 \pm 0.003$ & $0.991 \pm 0.007$ & $0.992 \pm 0.002$ & $0.985 \pm 0.012$ &  $0.984 \pm 0.005$ \\
\quad Other-Language Subspace & $0.989 \pm 0.006$ & $0.985 \pm 0.017$ & $0.988 \pm 0.009$ &  $0.935 \pm 0.127$ & $0.979 \pm 0.010$ \\
\quad Independent-Model Subspace & $0.001 \pm 0.002$ & $0.001 \pm 0.002$ & $0.004 \pm 0.016$ & $0.009 \pm 0.025$ &  $0.229 \pm 0.236$ \\
\quad Random Subspace & $0.001 \pm 0.002$ & $0.001 \pm 0.004$ & $0.002 \pm 0.011$ & $0.007 \pm 0.012$ &  $0.234 \pm 0.214$ \\
\bottomrule
\end{tabular}
}
\caption{Rank-8 answer-subspace ablations. We report the mean held-out
accuracy drop \(\pm\) standard deviation across 20 models, after averaging over
both language directions within each model. Random-subspace results are
additionally averaged over five independently sampled rank-matched subspaces.
Other-language subspaces are estimated from the other language in the same
bilingual model; independent-model subspaces are taken from a separately
initialized model.}
\label{tab:appendix-subspace-ablation}
\end{table*}

\begin{table*}[t]
\centering
\small
\setlength{\tabcolsep}{4pt}
\begin{tabular}{@{}lccccc@{}}
\toprule
\textbf{Condition} & \textbf{$r=2$} & \textbf{$r=4$} & \textbf{$r=8$} & \textbf{$r=16$} & \textbf{$r=32$} \\
\midrule
\multicolumn{6}{@{}l}{\textbf{Final-Residual Answer-Subspace Overlap}} \\
\quad \addonly{} & $0.577 \pm 0.250$ & $0.833 \pm 0.125$ & $0.944 \pm 0.080$ & $0.932 \pm 0.038$ & $0.931 \pm 0.030$ \\
\quad \addcopy{} & $0.636 \pm 0.293$ & $0.839 \pm 0.153$ & $0.928 \pm 0.091$ & $0.921 \pm 0.060$ & $0.933 \pm 0.032$ \\
\quad \subonly{} & $0.671 \pm 0.251$ & $0.810 \pm 0.153$ & $0.889 \pm 0.086$ & $0.863 \pm 0.063$ & $0.940 \pm 0.047$ \\
\quad \addsub{} & $0.833 \pm 0.255$ & $0.827 \pm 0.211$ & $0.855 \pm 0.176$ & $0.866 \pm 0.063$ & $0.956 \pm 0.029$ \\
\quad \addmax{} & $0.969 \pm 0.113$ & $0.878 \pm 0.126$ & $0.924 \pm 0.086$ & $0.947 \pm 0.032$ & $0.939 \pm 0.029$ \\
\addlinespace[3pt]
\multicolumn{6}{@{}l}{\textbf{Cross-Ablation Drop, Pre-MLP}} \\
\quad \addonly{} & $0.296 \pm 0.156$ & $0.601 \pm 0.144$ & $0.928 \pm 0.036$ & $0.993 \pm 0.002$ & $0.993 \pm 0.003$ \\
\quad \addcopy{} & $0.238 \pm 0.138$ & $0.531 \pm 0.152$ & $0.923 \pm 0.046$ & $0.991 \pm 0.004$ & $0.993 \pm 0.002$ \\
\quad \subonly{} & $0.497 \pm 0.200$ & $0.754 \pm 0.126$ & $0.956 \pm 0.046$ & $0.994 \pm 0.003$ & $0.994 \pm 0.003$ \\
\quad \addsub{} & $0.485 \pm 0.169$ & $0.752 \pm 0.126$ & $0.947 \pm 0.047$ & $0.990 \pm 0.005$ & $0.991 \pm 0.004$ \\
\quad \addmax{} & $0.835 \pm 0.112$ & $0.958 \pm 0.027$ & $0.983 \pm 0.005$ & $0.985 \pm 0.004$ & $0.985 \pm 0.004$ \\
\addlinespace[3pt]
\multicolumn{6}{@{}l}{\textbf{Cross-Ablation Drop, Final}} \\
\quad \addonly{} & $0.564 \pm 0.179$ & $0.874 \pm 0.080$ & $0.989 \pm 0.006$ & $0.992 \pm 0.003$ & $0.993 \pm 0.002$ \\
\quad \addcopy{} & $0.550 \pm 0.193$ & $0.861 \pm 0.082$ & $0.985 \pm 0.017$ & $0.990 \pm 0.009$ & $0.993 \pm 0.003$ \\
\quad \subonly{} & $0.724 \pm 0.120$ & $0.940 \pm 0.029$ & $0.988 \pm 0.009$ & $0.989 \pm 0.008$ & $0.991 \pm 0.007$ \\
\quad \addsub{} & $0.668 \pm 0.230$ & $0.824 \pm 0.231$ & $0.935 \pm 0.127$ & $0.983 \pm 0.016$ & $0.989 \pm 0.005$ \\
\quad \addmax{} & $0.858 \pm 0.087$ & $0.948 \pm 0.027$ & $0.979 \pm 0.010$ & $0.983 \pm 0.006$ & $0.985 \pm 0.005$ \\
\bottomrule
\end{tabular}
\caption{Rank sensitivity of answer-subspace overlap and causal
cross-language ablation. Cells report means $\pm$ standard deviations
across the same 20 frozen models per condition. Subspace bases are fitted
on training examples; ablation drops are evaluated on held-out examples
and averaged over both language directions within each model. Rank 8 is
the primary analysis. Larger ranks mechanically span and remove more
directions, so magnitudes should not be compared across ranks as though
rank were an experimental condition. The sweep tests whether the qualitative
same-model overlap and cross-language causal dependence hinge on the primary
rank choice. Rank-8 random- and independent-model controls are reported in
\Cref{tab:appendix-subspace-overlap-controls,tab:appendix-subspace-ablation}.}
\label{tab:appendix-rank-sensitivity}
\end{table*}

\end{document}